\documentclass{article}
\usepackage[utf8]{inputenc}
\usepackage{amsmath}

\title{Conditions for Convergence in Regularized Machine Learning Objectives}
\author{Patrick Hop\\Mathematics, UC Berkeley \and Xinghao Pan\\EECS, UC Berkeley}
\date{May 16th, 2013}

\usepackage{graphicx}
\usepackage[margin=0.75in]{geometry}
\usepackage{amsfonts}

\begin{document}

\maketitle

\section{Abstract}
    Analysis of the convergence rates of modern convex optimization algorithms can be achived through binary means: analysis of emperical convergence, or analysis of theoretical convergence. These two pathways of capturing information diverge in efficacy when moving to the world of distributed computing, due to the introduction of non-intuitive, non-linear slowdowns associated with broadcasting, and in some cases, gathering operations. Despite these nuances in the rates of convergence, we can still show the existence of convergence, and lower bounds for the rates. This paper will serve as a helpful cheat-sheet for machine learning practitioners encountering this problem class in the field.
    
\section{Primal Structure}
Consider the following primal optimization problem, where $\ell$ and $r$ are both convex, and $r$ can be non-smooth.

~

$\underset{~x\in \mathbb{R}^n}{\min}$ $\ell(x) + r(x)$

~

This structure, that we will assume, encapsulates problems such as the least absolute selection and shrinkage operator (LASSO), where $\ell$ is smooth and $r$ is non-smooth, and support vector machines (SVMs), where both $\ell$ and $r$ are smooth \cite{boyd2013}.


\section{Convergence Conditions}
The following six convergence conditions are necessary and sufficient.

\subsection{$\ell$ is lipschitz continuous}
    Lipschitz continuity bounds how fast fast a continous function can change: $\forall$ points on the graph of a lipschitz continuous function, the absolute value of the slope of the line connecting these two points is bounded by some definte-real number, called it's lipschitz constant.
    The implication of this is that when descending a lipschitz continuous function, we no longer have to consider violent fluctuations in the gradient; this will turn out to be a valuable inference when selecting stepsizes later on.

\subsubsection{lipschitz continuity: assuming $\ell$ is convex, and $\ell '$ is L-Lipschitz Continuous}

~~~ $||\ell'(x) - \ell'(y)|| \le L||x - y||$ ~~ $\forall$ $x,y$ $\in$ $\mathbb{R}^n$ \cite{schmidt2011}
    
\subsubsection{lipschitz continuity: assuming twice-differentiability}
I.E the eigenvalues of the Hessian are bounded above by L \cite{schmidt2011}.

~

$0 \preceq \ell''(x) \preceq LI$ ~~ $\forall ~ x \in \mathbb{R}^n $

\subsection{$r$ is a lower semi-continuous proper convex function}
For some pre-image $x$, image values for nearby $x$ are near the image of $x$, or less than the image of $x$. This is a weaker notion of continuity for extended-real valued functions \cite{schmidt2011}.

\begin{figure}[h!]
\centering
\includegraphics[scale=.4]{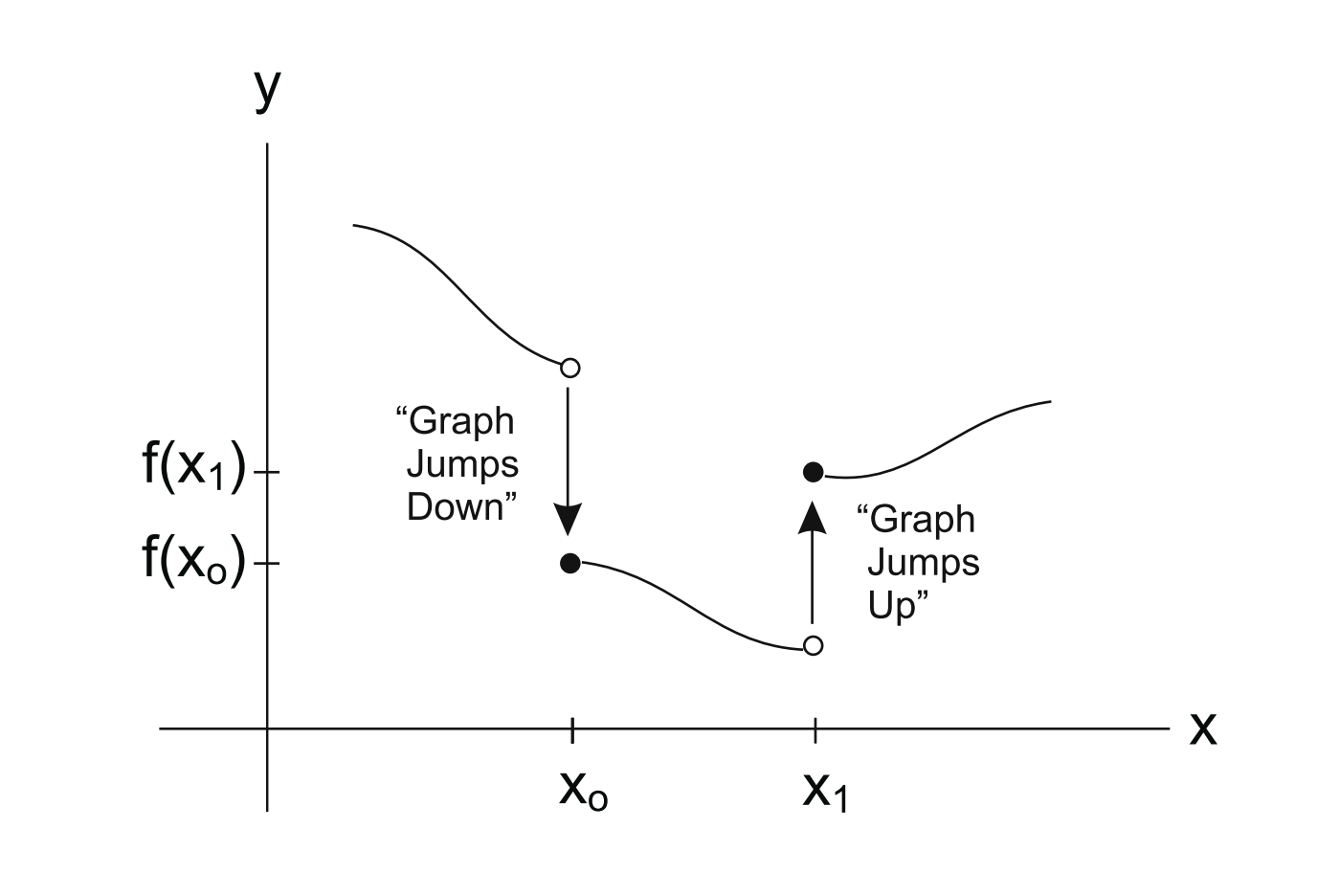}
\caption{A function that is LSC at $x_0$, but not at $x_1$}
\end{figure}

\subsection{$\ell + r$ attains its min at a certain x*}
That is, a minimum exists; this shouldn't be a surprise for convex regularizers, as the sum of two convex functions is too, convex \cite{schmidt2011}.

~

$p^* := \ell(x^*) + r(x^*)$

\subsection{the stepsize $\alpha _ k$ is set to 1/L}
~~~ $\alpha_k := 1/L$ \cite{schmidt2011}

\subsection{the gradient $\ell'$ is computed with a error $\epsilon _k$}
The consequence of this condition is that we can, indeed, compute our gradient \cite{schmidt2011}; this sometimes isn't the case for some more exotic problems \cite{boyd2013}.

~

$\ell^{'*} = \ell^{'} + \epsilon_k $

\subsection{$x_k$ is an $\epsilon _k$-approximate solution of the proximity operator}
Like the conditions on the gradient, this condition, when satisfied, means that we can compute our proximal, subject to some error $\epsilon_k$ \cite{schmidt2011}.

~

$x^*_k = x_k + \epsilon_k = \underset{~y\in R^n}{argmin}$ $f(y) + 1/2||x-y||^2$

\section{Error Conditions for Inexact Methods}
Often times it is purely impossible, or computationally unatractive, to compute an exact gradient or proximal; in these cases, it's neccessary to analyse the behavior of the sequence of errors in the limit; this is the case for the gradient, when using the mini-batch method for a trivially seperable $\ell$ \cite{cotter2011}.

Provided are the conditions on the sequences of gradient errors, $e_k$, and sequences of proximity errors $\epsilon_k$ for a selection of the aforementioned methods.

\subsection{Proximal-Gradient (Convex)}
For the basic Prox-Grad where $\ell$ is convex, the sequence of the normed errors of the gradient, ${||e_k||}$, and the sequence of errors of the proximal, ${\sqrt{\epsilon_k}}$, are summable and decrease as $O(1/k^{1+\delta})$ for any $\delta > 0$ \cite{schmidt2011}.

\subsection{Proximal-Gradient (Strongly Convex)}
${||e_k||}$ and ${\sqrt{\epsilon_k}}$ must decrease to zero linearly \cite{schmidt2011}.

\subsection{Accelerated Proximal-Gradient (Convex)}
For accelerated Prox-Grad where $\ell$ is convex, the sequence of the normed errors of the gradient, ${||e_k||}$, and the sequence of errors of the proximal, ${\sqrt{\epsilon_k}}$, are summable decrease as $O(1/k^{2+\delta})$ for any $\delta > 0$ \cite{schmidt2011}.

\subsection{Accelerated Proximal-Gradient (Strongly Convex)}
${||e_k||^2}$ and ${\epsilon_k}$ must decrease linearly to zero \cite{schmidt2011}.

\section{Convergence Rates}

\begin{table}[ht]
\centering 
\begin{tabular}{c c c c} 
\multicolumn{4}{c}{Note: Stochastic Methods include an additive $\sigma/\sqrt{k}$ term \cite{cotter2011}}\\
\hline\hline 
 & Convex & Strongly Convex \\ [0.5ex] 
\hline
Sub-Gradient & $O(1/\sqrt{k})$ & $O(1/k)$ & \\
Prox-Gradient & $O(1/k)$ & $O((1-\mu / L)^k)$ &\\
Accelerated Prox-Grad & $O(1/k^2)$ & $O((1-\sqrt{\mu / L})^k)$\\
ADMM & $O(1/k)$ & $n/a$ &\\
[1ex] 
\hline 
\end{tabular}
\caption{Convergence Rates \cite{schmidt2011}, \cite{heYuan2011}} 
\label{table:nonlin} 
\end{table}

\section{Conclusions and Further Research}
In summary, there is a substantial amount of mathematical theory backing the convergence conditions and rates of the proximal methods utilized to solve regularized machine learning objectives. Luckily, a substantial subset of modern machine learning algorithms, such as support vector machines (SVMs), and the least absolute selection and shrinkage operator (LASSO) \cite{boyd2013}, can be casted into this form, making the outlined theory in this short paper sufficient for practical use by machine learning practioners in the field.

Further research will include insight into the convergence rate of the strongly convex case of ADMM, and perhaps even an investigation into the theory behind the vexing emperical results that emerge when objective seperability is exploited, and the computation is distributed over n-machines.

\end{document}